%% file: main.tex
\documentclass[letterpaper, 10 pt, conference]{ieeeconf}  

\IEEEoverridecommandlockouts                              

\usepackage{url}
\usepackage{pifont}
\usepackage{graphicx} 
\usepackage{xcolor} 
\usepackage{booktabs}
\usepackage{amssymb}    
\usepackage{amsmath}
\usepackage{subcaption}
\usepackage{float}
\usepackage{multirow}
\usepackage{todonotes}
\usepackage{caption}
\usepackage{hyperref}
\usepackage{flushend}

\newcommand{\rot}[1]{\rotatebox{90}{#1}}
\newcommand{\xmark}{\ding{55}}%

\title{\LARGE \bf
VPR-Space: A Benchmark for Visual Place Recognition\\in Space Analog Domains
}

\author{Richard Fechner$^{1}$, Peer Neubert$^{1}$
\thanks{$^{1}$Institute for Computational Visualistics, University of Koblenz} 
\thanks{correspondence: \texttt{rfechner@uni-koblenz.de}}
}

\begin{document}

\maketitle
\pagestyle{plain}

\begin{abstract}
	
Visual Place Recognition (VPR) is an important means for loop closure detection in visual Simultaneous Localization and Mapping (SLAM).
While state-of-the-art VPR and SLAM approaches show impressive results in urban and feature-rich natural environments, the vast, repetitive, and feature-poor environments of extraterrestrial surfaces like on the Moon or Mars, together with their challenging illumination conditions, pose severe challenges for  existing approaches. This paper presents a novel benchmark, VPR-Space, to evaluate VPR capabilities on extraterrestrial surfaces.
There are three main contributions:
(1) We collected and curated sixteen VPR database-query datasets based on existing data from Space Analog Domains (SAD), terrestrial environments that resemble properties of the Moon or Mars. The collection includes a wide range of challenges regarding environment, viewpoint change, and illumination conditions. All datasets were manually corrected for ground-truth camera pose errors and unwanted camera artfacts.
(2) To address the ill-posed nature of VPR, we propose a novel approach to generate the ground-truth information in VPR benchmarks. 
Instead of defining that a query and datbase image show the same place if and only if their camera poses are within fixed position and viewing direction thresholds, we propose to make the decision dependent on whether the match would be valuable in the downstream SLAM task.
To avoid the overhead and restrictions of evaluating VPR in-the-loop within SLAM systems, we provide VPR ground-truth (GT) matrices based on the performance of an ensemble of state-of-the-art Relative Pose Estimators (RPE). 
Under this GT we require a VPR method to identify exactly those image pairs for which a subsequently applied RPE is expected to provide a valuable pose estimate in a SLAM context.
(3) We evaluated nine state-of-the-art VPR approaches. The results show considerable room for improvement and indicate directions for future research. 
Detailed results, data, and code will be made available upon publication of the paper. We provide a pre-release version of the dataset on github: \texttt{\url {https://github.com/rfechner/VPR-Space}}

\end{abstract}
\section{Introduction}

\begin{figure}[t]
	\centering
	\includegraphics[width=\linewidth]{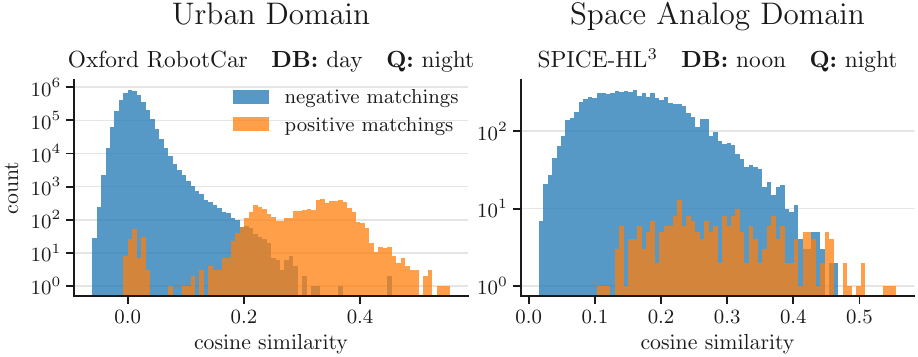}
	\caption{Image descriptor similarity histograms for Oxford RobotCar day--night \cite{maddern20171} (left) and SPICE-HL³ noon--night \cite{rodriguez2026spice} (right). We used Bag-of-Queries \cite{ali2024boq}, a SotA VPR method, to compute the descriptors. The distributions in the space analog domain are even less separable than in this challenging urban domain (i.e., there is no similarity threshold which approximately separates the positive and negative matchings).}
	\label{fig:challenges_space}
	\vspace{-0.5cm}
\end{figure}

Reliable camera-based navigation is an important capability of autonomous mobile systems in GNSS-denied environments such as the surfaces of the Moon and Mars. The potential of, and challenges associated with, such navigation systems have recently been highlighted by the initial success and subsequent crash of the Mars helicopter Ingenuity. A likely cause being the inability of the visual navigation system to find features on the homogeneous surface of Mars~\cite{jpl2024ingenuity}.
In the context of camera-based navigation, Visual Place Recognition (VPR) addresses the problem of identifying  matching image pairs between a query set and a database set of images depicting the same places. It is an important component of loop-closure detection in Simultaneous Localization and Mapping (SLAM) where it enables the correction of accumulated errors of odometry-based trajectory estimates.
Existing SLAM and VPR approaches achieve impressive performance in urban and feature-rich natural environments like forests. However, the visual characteristics of extraterrestrial planetary surfaces, such as those of the Moon or Mars, differ fundamentally from these environments. They more closely resemble Earth's stone and sand deserts or volcanic sites, exhibiting regolith and topographic structures such as craters and wrinkle ridges. The vast, often homogeneous and feature-poor landscapes, together with harsh illumination conditions can pose severe challenges for image descriptors. 
This is illustrated in Fig.~\ref{fig:challenges_space}, which shows the distributions of descriptor similarities for image pairs depicting either the same place (orange) or different places (blue). The two distributions are even more difficult to separate in the challenging Moon-like environment (right) than in the challenging urban environment (left). 
Combined with the low error tolerance of space missions, this motivates the need for careful analysis and evaluation of visual navigation algorithms under planetary-like conditions. 

\begin{figure*}[t]
	\centering
	\begin{minipage}[t]{0.48\textwidth}
		\centering
		\includegraphics[width=\textwidth]{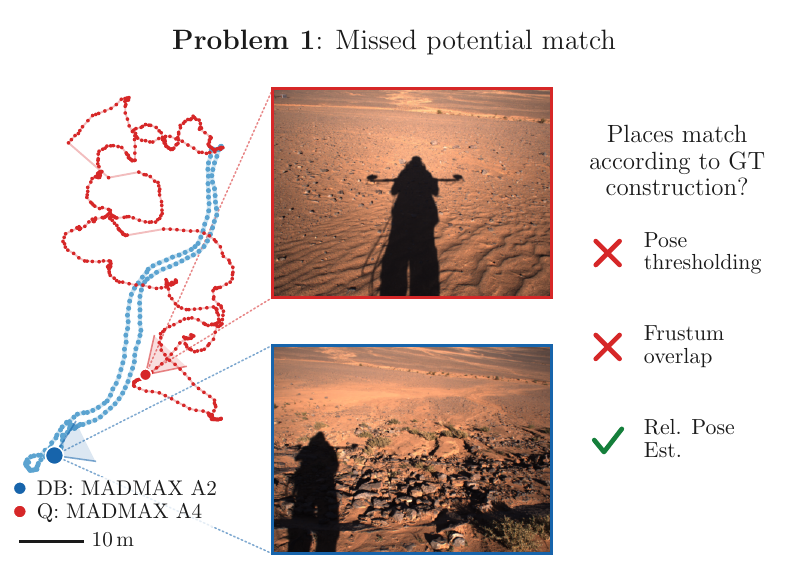}
		\captionof{figure}{Two images of the MADMAX A2--A4 dataset, captured 20m apart. Constructing the VPR-GT by pose thresholding or frustum overlap fails to capture this match, missing out on a potential match. RPEs however are able to estimate a low error relative pose change, making this a desired VPR match.}
		\label{fig:rep-advantages:a}
	\end{minipage}%
	\hfill%
	\begin{minipage}[t]{0.48\textwidth}
		\centering
		\includegraphics[width=\textwidth]{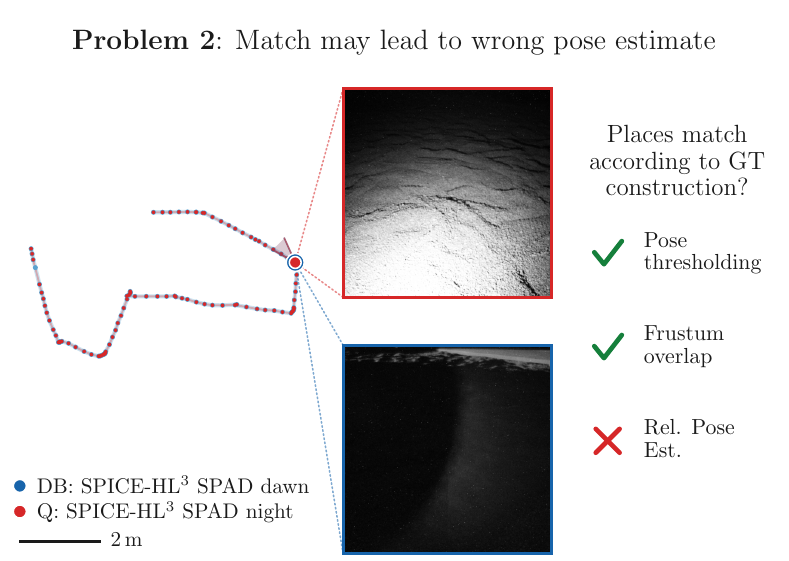}
		\captionof{figure}{Two images of the SPICE-HL³ SPAD dawn--night dataset, captured at the same location but with different lighting conditions. Based on pose thresholding or frustum overlap these images should match. Employing a set of SotA RPEs, the minimum rotation error is $> 31^\circ$; matching both images is unlikely to result in a correct pose estimate. We argue to label this pair a non-match.}
		\label{fig:rep-advantages:b}
	\end{minipage}
	\vspace{-0.5cm}
\end{figure*}

In this paper, we present VPR-Space, a benchmark for VPR approaches that provides (1) challenging combinations of curated database and query image sets with planetary-like conditions, and (2) carefully created ground-truth information about which image pairs depict the same place. Both parts are non trivial.
Collecting large amounts of extraterrestrial on-site data is impractical. An established alternative for preparation of space missions is using Space Analog Domains (SADs), areas on the Earth that resemble properties of the Moon or Mars~\cite{catalogue}. Table~\ref{tab:overview} lists available SAD-based datasets for the evaluation of SLAM systems.
Sec.~\ref{sec:approach} will describe how we selected, manually curated, and preprocessed the available data in order to create representative database and query sets for the VPR-Space benchmark.

For the second part, creating the ground-truth (GT) information for evaluation of VPR approaches, a major challenge is the ill-posed nature of the VPR problem: There is no unique correct answer whether two images show the same place or not~\cite{schubert2021makes}. 
As will be discussed in Sec.~\ref{relwork:vpr-gts}, the usual approach to create GT in existing VPR datasets is based on the ground truth pose difference between the two images (\textit{Pose thresholding}) or overlap in field of view with a limited maximum distance (\textit{Frustum overlap}). 
In this paper, we propose a different approach to VPR-GT generation: According to \cite{garg2021your}, the quality of a VPR result is driven by its value for the downstream task. To avoid the overhead and restrictions of evaluating VPR in-the-loop within a SLAM system, we propose to use the performance of an ensemble of Relative Pose Estimators (RPE) as an oracle that identifies valuable matches (i.e., the RPE quality is used to generate the GT matrices). This can address two problems of VPR-GT generation in large and open environments, and in environments with very harsh illumination conditions, both common properties of SADs: 

\begin{itemize}
	\item \textbf{Problem 1}: Fig.~\ref{fig:rep-advantages:a} illustrates how two images that can be considered to show the same place but which are taken far apart from one another, do not match when deciding correspondence based on pose distance or frustum overlap. The proposed RPE-GT includes this match if and only if we expect RPE approaches to find a valuable pose transformation.

	\item \textbf{Problem 2}: Fig.~\ref{fig:rep-advantages:b} illustrates an example image pair that is taken from the exact same camera pose but due to extreme lighting conditions, a low error relative pose estimate cannot be made. A GT based on pose distance or frustum overlap would include this match, although it would potentially introduce additional errors in a SLAM system.
	
\end{itemize}

Sec.~\ref{sec:approach_gt} will provide details on this RPE-based ground truth creation.
Sec.~\ref{sec:results} will then evaluate state-of-the-art (SotA) VPR approaches on the proposed benchmark.
In summary, the main contributions of this paper are:

\begin{enumerate}
    \item \emph{VPR-Space}, a challenging VPR benchmark for space analog domains, with 16 database-query sequence pairs derived from existing space analog SLAM datasets.
    \item We propose and discuss a novel method to construct an application-driven ground truth for VPR. It leverages a set of SotA relative pose estimators to decide whether two images should match.
    \item We evaluate nine SotA VPR methods on \emph{VPR-Space}.
\end{enumerate}

The benchmark with detailed results, evaluation code, as well as the code to reconstruct the benchmark from the raw SLAM datasets will be made available on the benchmark website upon paper publication.

\section{Related Work}
\label{sec:rel_work}


\subsection{VPR Datasets and Benchmarks} 
VPR is an intensively studied problem and the literature discusses typical challenges such as viewpoint changes (observing places from different camera poses) and appearance changes (e.g., due to varying illumination, weather, or season)~\cite{schubert2021makes, garg2021your, schubert2023visual}. To successfully perform VPR, extracted features must 1) be present in the environment, and 2) be processable (i.e., observable, extractable, describable, and matchable).
Existing VPR datasets cover a broad spectrum of environments and visual dynamics. Datasets like Oxford RobotCar \cite{maddern20171} test VPR in urban domains, capturing the streets of Oxford in different lighting and environmental conditions. The Tokyo 24/7 \cite{torii201524} dataset shows views of the streets of Tokyo at day and night, additionally introducing the challenge of changing environments. Semi-structured VPR datasets include Nordland \cite{nordland}, a railway trip through Norway under changing seasons, or WildCross \cite{knights2026wildcross}, a recent dataset addressing the gap of cross-modal VPR in natural domains. To evaluate general performance, VPR benchmarks aggregate these distinct domains. For example,  VPR-Bench~\cite{zaffar2021vpr} combines a mix of indoor and outdoor (urban and semi-structured) VPR datasets to provide a balanced assessment across varied scenarios. However, existing benchmarks predominantly reflect terrestrial features (e.g., roads, vegetation, and man-made structures). 

\subsection{SLAM Space Datasets}\label{relwork:spacedatasets} 
Although, to the best of our knowledge, no dedicated SAD VPR datasets are available, Table~\ref{tab:overview} provides properties of existing SLAM Space datasets that will be used for VPR benchmark construction in Sec.~\ref{sec:approach}.
 We want to highlight four particularly relevant datasets. The MADMAX dataset \cite{meyer2021madmax} captures 36 trajectories on 8 different sites near the city of Erfoud, Morocco. The collected trajectories depict dunes, coarse stone desert and traverses along and over rock-structures. The S3LI Etna dataset \cite{giubilato2022challenges} was collected around the crest and in close proximity of Mt. Etna, Italy. The sequences collected along the crater ridge depict grayscale imagery of volcanic rock and ash, a very monotonous landscape. The LunarLoc dataset \cite{thomas2026lunarloc} publishes 17 sequences along a 27m by 27m simulated lunar landscape, consisting of images collected with a digital twin of the NASA IPEx rover. The recently published SPICE-HL³ dataset \cite{rodriguez2026spice} was collected at the LunarLab in Luxembourg. It includes a particularly relevant traverse that follows a stop-and-capture routine with several sequences of the same fixed trajectory, each sequence providing differently illuminated images. 

\subsection{Approaches to Construct a VPR-GT}\label{relwork:vpr-gts} 
Constructing the VPR-GT requires the critical decision of which images \emph{should} match in the first place, a topic discussed at length by Schubert et al. \cite{schubert2021makes}. Adjacently, Garg et al. \cite{garg2021your} name ``drivers" as the motivators for forming the VPR-GT.
The established VPR benchmarks usually construct their VPR ground truths based on the distance in poses between two images or match individual images based on frame index. For example, the Tokyo 24/7 dataset \cite{torii201524} decides correspondence based on GPS distance of collected images; a found database image is a positive if it was captured at most 25m away from the query image. The Nordland dataset \cite{nordland} matches image pairs based on their frame index, as database and query pairs are collected along the same trajectory and time-synchronized. A recent work on benthic VPR \cite{Larsen_2026} uses an evaluation that is closer to the downstream task. The approach estimates 3D footprints on the seafloor and determines a top-down feature overlap to decide whether two images correspond to the same VPR-place. In a similar direction, the authors of the S3LI Vulcano dataset \cite{giubilato2026s3livulcanodatasetdataset} mention the possibility to calculate VPR ground truths from point-cloud overlaps, when LiDAR or precise depth maps are available. They describe a VPR-GT construction built on shared visible information in the frustum but leave the construction of a decision rule to future work.


\begin{table}[ht]
\centering\footnotesize
\caption{Sensor modalities relevant to VPR and attributes of SLAM datasets in space domains. The properties Viewpoint Change (VC) and Illumination Change (IC) denote the existence of hard viewpoint- and illumination changes within and between two sequences of the dataset respectively.}
\label{tab:overview}
\begin{tabular}{@{}llllllllll@{}}
\toprule
Dataset  & \rot{Grayscale} & \rot{RGB} & \rot{LiDAR} & \rot{IMU/Gyro} & \rot{GNSS/Pos} & \rot{hard VC} & \rot{hard IC} & \rot{Mars analog} & \rot{Moon analog} \\
\midrule
Devon Island \cite{furgale2012devon}           & \xmark & \checkmark &  \xmark & \xmark & \checkmark & \checkmark & \xmark & \checkmark & \xmark\\
MADMAX \cite{meyer2021madmax}                  & \xmark & \checkmark& \xmark & \checkmark & \checkmark & \checkmark & \xmark & \checkmark & \xmark\\
S3LI Vulcano \cite{giubilato2026s3livulcanodatasetdataset}  & \xmark & \checkmark & \checkmark & \xmark & \checkmark & \checkmark & \xmark & \checkmark & \xmark\\
S3LI Etna \cite{giubilato2022challenges}  & \checkmark &\xmark & \checkmark & \checkmark & \checkmark & \checkmark & \xmark & \checkmark & \xmark\\
Erfoud  \cite{lacroix2020erfoud} & \checkmark & \xmark &  \checkmark & \checkmark & \checkmark & \checkmark & \xmark & \checkmark & \xmark\\ 
ROBEX  \cite{vayugundla2018datasets} & \checkmark & \xmark & \xmark & \checkmark & \checkmark & \checkmark & \xmark & \checkmark & \checkmark\\
BASEPROD \cite{gerdes2024baseprod} &\xmark & \checkmark & \xmark & \checkmark & \checkmark & \checkmark & \xmark & \checkmark & \xmark\\
ENAV  \cite{lamarre2020canadian} &\checkmark & \xmark & \xmark & \checkmark & \checkmark & \checkmark & \xmark & \checkmark & \xmark\\
Katwijk  \cite{hewitt2018katwijk} & \xmark & \checkmark & \checkmark & \checkmark & \checkmark & \checkmark & \xmark & \checkmark & \xmark\\
LunarLoc \cite{thomas2026lunarloc} & \checkmark & \xmark & \xmark & \checkmark & \checkmark & \checkmark & \xmark & \xmark & \checkmark\\
SPICE-HL³ \cite{rodriguez2026spice} & \checkmark & \xmark & \xmark & \checkmark & \checkmark & \checkmark & \checkmark & \xmark & \checkmark\\
POLAR Traverse \cite{hansen2024polar} & \checkmark & \xmark & \xmark & \xmark & \xmark & \checkmark & \checkmark & \xmark & \checkmark\\ 
TAILPlus \cite{wang2024readyplanetaryexplorationrobots} & \xmark & \checkmark & \checkmark & \checkmark & \checkmark & \checkmark & \checkmark & \checkmark & \xmark\\
\bottomrule
\end{tabular}
\vspace{-0.3cm}
\end{table}


\section{Benchmark Construction}
\label{sec:approach}

A VPR benchmark for SADs should include a diverse set of challenging scenarios and environments which can be used to gain a holistic understanding of the capabilities of a VPR method in SADs. We see four major design requirements for such a benchmark: (1)~It should resemble relevant and established VPR challenges like Viewpoint Change (VC) and Illumination Change (IC) \cite{schubert2023visual}. (2) It should include a diverse set of space analog environments and topologies, representing both, Mars- as well as Moon-analog domains. (3) It should have an application relevant VPR-GT which addresses the specific challenges of the domain. (4) It should challenge the SotA for VPR. In the following, we will show how \emph{VPR-Space} meets these requirements.

\begin{figure*}
  \includegraphics[width=\textwidth]{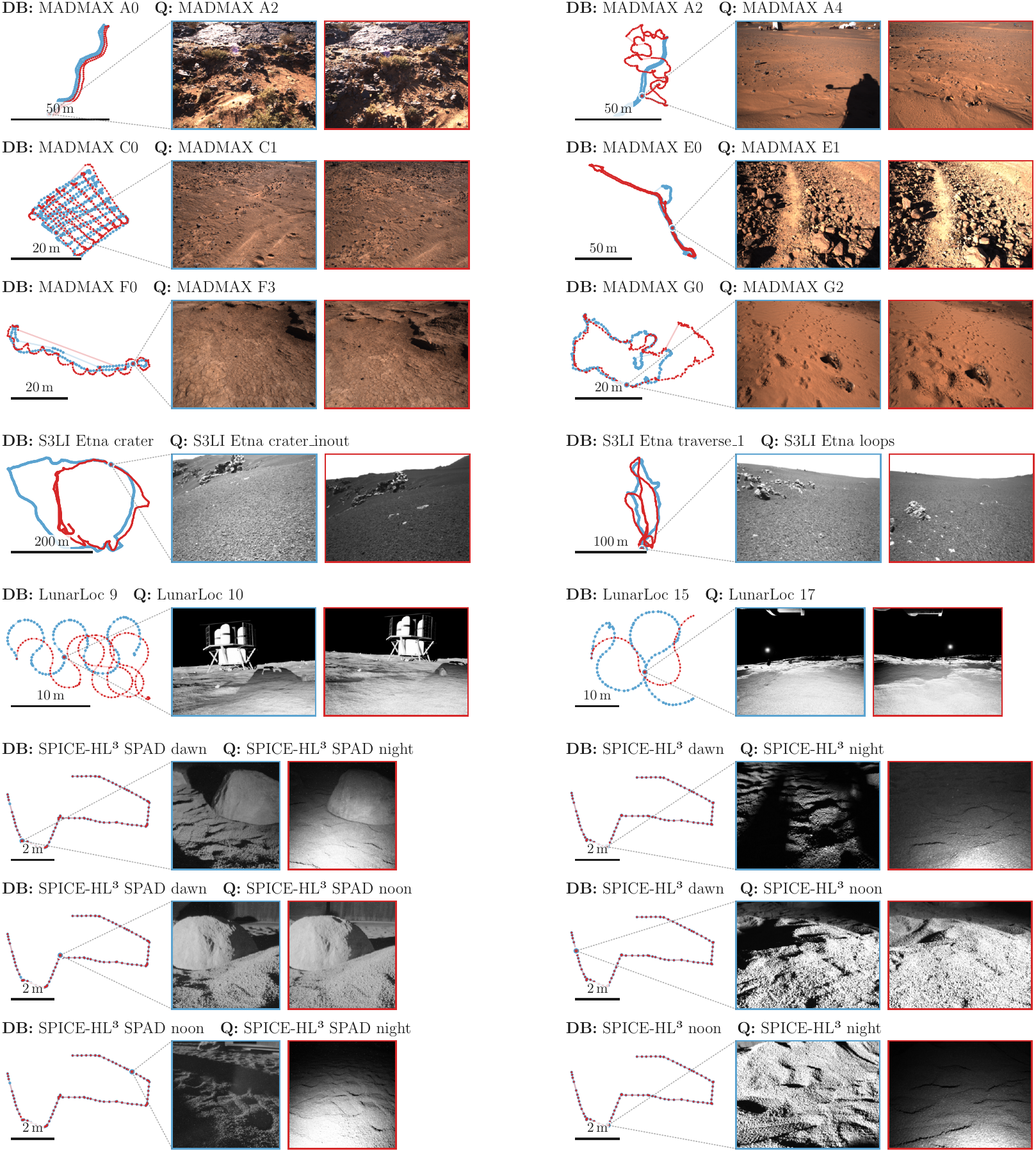}
  \caption{An overview of the database (blue) and query (red) sequences of the VPR-Space benchmark. We show example matches from both sequences to illustrate the shape of the task. For SPICE-HL³, database and query sequence lie on top of each other.}
  \label{fig:bench-overview}
\end{figure*}

\subsection{Choosing Database and Query Sequences}
From the available datasets in Table~\ref{tab:overview}, we select 16 sequence pairs in total. Trajectory shapes and example images can be seen in Figure \ref{fig:bench-overview}. We identify ten pairs of diverse and challenging database and query sequences falling into the VC category -- six from MADMAX and two from S3LI and LunarLoc respectively. These pairs contain changing viewpoints of the same place. They are distinct sequences collected at the same site with crossover points and intervals. For the VPR challenge of IC we chose six pairs of sequences from SPICE-HL³'s A trajectory. The sequences follow a fixed trajectory but differ in lighting, camera sensor and exposure time. For the two cameras, one monochrome and one Single Photon Avalanche Diode (SPAD), we chose three main setups: dawn--night, noon--night and dawn--noon.  
We apply a sub-sampling routine to achieve an even distribution of data points across the trajectories. Details can be found in the Appendix.
Overall, this selection satisfies our design requirements (1) and (2), as it consists of 10 VC and 6 IC setups, balances Mars- and Moon-analogs, has different trajectory topologies and geological environments.

\subsection{Data Cleaning}


Our VPR-GT construction requires correct ground-truth camera poses in order to evaluate the result of an RPE. During initial data analysis, we noticed inconsistencies in the provided poses and imagery and decided to manually verify the selected sequences. In MADMAX and S3LI Etna we encountered GNSS and IMU errors, in the SPICE-HL³ sequences we observed some camera artifacts. We verified the data with three independent human annotators, using a shared labeling UI where position, orientation and associated image are visible. Any data point being (a) visually corrupted or (b) exceeding an apparent GT pose error beyond 1 meter in position or 10° in orientation was flagged. Out of about 9200 flagged data points, the three annotators agree on about 72\%; at least two annotators agree on 81\% of the flagged data points. We discarded every frame flagged by at least one annotator. The found errors are provided as part of the published data, so that they can be reused by others working with the raw data. 

\subsection{Construction of the VPR Ground Truth}
\label{sec:approach_gt}

The VPR ground truth is usually represented by discrete sets of matching image pairs from query and database sets, e.g., in form of a ground-truth matrix~\cite{schubert2023visual}. 
We propose to create this matrix using an ensemble of available SotA Relative Pose Estimator approaches that serve as an oracle to tell whether a valuable pose estimate could be obtained from an image pair in a SLAM downstream task. 
The RPE ensemble is used in combination with ground-truth pose information. Instead of directly judging the ground-truth pose distance between a query and a database image, we evaluate the error between the output of the RPEs and the ground-truth pose transformation to decide whether an image pair is valuable in a SLAM context (and thus should be found by an VPR approach).
We want to emphasize that, in contrast to an actual end-to-end evaluation in a SLAM system, this oracle has to be used only once for each image pair and not for each evaluated VPR approach.


We construct a majority voting scheme over multiple RPEs. We leverage off-the-shelf monocular metric-relative pose estimation pipelines that predict a 6DoF relative pose consisting of rotation and scaled translation. 
Where applicable, backbone architectures are provided with metric depth predictions from Depth Anything V2 \cite{depth_anything_v2} and refined using RANSAC \cite{fischler1981random}. Our evaluation ensemble comprises DUSt3R \cite{dust3r_cvpr24}, MUSt3R \cite{must3r_cvpr25}, MapAnything \cite{keetha2026mapanything}, MADPose \cite{yu2025madpose}, and a SuperPoint \cite{detone2018superpoint} + LightGlue \cite{lindenberger2023lightglue} combination.


\begin{figure}[h!]
    \centering
    \includegraphics[width=0.98\linewidth]{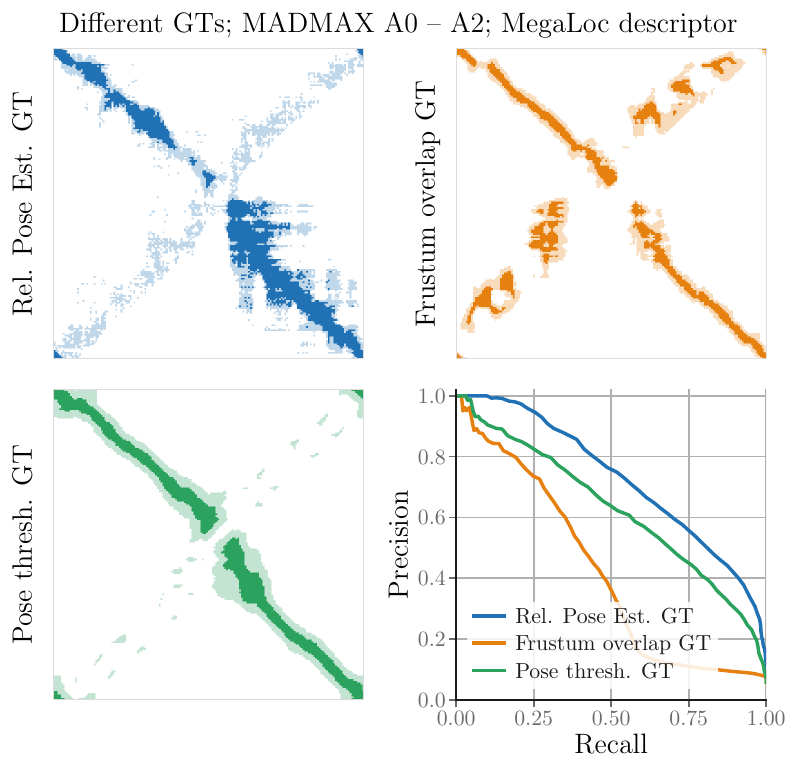}
    \vspace{-0.2cm}
    \caption{GT matrices (hard GT in dark, soft GT in light color) and Precision-Recall Curves for VPR-GTs of different constructions. To enable comparison, the number of positive matches is equalized across constructions. 
	This example illustrates that established GT constructions tend to request harder matches. This pattern repeats across the dataset.
	}
	\vspace{-0.2cm}
    \label{fig:prcurves}
    \vspace{-0.4cm}
\end{figure}

Two images $I_1$ and $I_2$ become a match in the GT matrix if a sufficient number of RPEs produces a small relative pose error between the two images. An RPE $f^{(i)}$ takes two images $I_1, I_2$, associated depth images $D_1, D_2$ and the camera intrinsics matrix $K \in \mathbb{R}^{3 \times 3}$. It outputs a relative rotation $\hat R^{(i)} \in SO(3)$ and translation $\hat t^{(i)} \in \mathbb R^{3}$. Since the ordering of input images can influence the results, each RPE produces two estimates: \mbox{$f^{(i)}(I_1, I_2, D_1, D_2, K) = \hat R_{1,2}^{(i)}, \hat t_{1,2}^{(i)}$} and \mbox{$f^{(i)}(I_2, I_1, D_2, D_1, K) = \hat R_{2,1}^{(i)}, \hat t_{2,1}^{(i)}$}. From known ground truth relative rotation $R_\text{GT}$ and translation $t_\text{GT}$ we calculate the geodesic rotation error $\Delta R^{(i)} \in [0, 180)$ [°] in degrees and the euclidean distance $\Delta t^{(i)} \in [0, \infty)$ [m] for the respective input ordering. In all experiments we relax $\Delta t^{(i)}$ to \mbox{$\Delta t^{(i)} := \min_s \|s \cdot \hat t^{(i)} - t_\text{GT}\|_2$}, the minimal distance to the true translation $t_\text{GT}$ on the span of the estimated translation direction $\hat t^{(i)}$. This reduces the required relative pose estimate to five degrees of freedom, making it easier for the RPEs to predict a correct translation. We say that RPE $f^{(i)}$ estimates the relative pose correctly if for both input orderings its corresponding estimate lands below fixed thresholds: \mbox{$\Delta R^{(i)}_{1,2}<\tau_R\wedge\Delta t^{(i)}_{1,2}<\tau_t$} and \mbox{$\Delta R^{(i)}_{2,1}<\tau_R\wedge\Delta t^{(i)}_{2,1}<\tau_t$}. We decide that the two images correspond if this condition holds for at least $n$ RPEs. Setting $n>1$ regularizes the RPE ensemble, reducing the risk to add image pairs where an RPE created an arbitrary transformation that is only correct by chance.
Additionally, images taken more than $d_\text{max}$ meters apart are never positives as we observed that RPE errors grow too large with the distance between input images. 

Building upon this majority voting mechanism, we can vary the error thresholds $\tau_R, \tau_t$, and $n$ to construct distinct soft- and hard- correspondence matrices required for a VPR evaluation. We follow the work of Schubert et al. \cite{schubert2023visual} where the soft ground truth denotes matches outside the hard ground truth which should not be penalized when an algorithm retrieves them. Exact hyperparameter choices and resulting dataset statistics are provided in the Appendix. Figure \ref{fig:prcurves} compares this approach to constructing the VPR-GT based on pose thresholding or frustum overlap.

\begin{figure*}[t]
	\centering
	\begin{subfigure}[t]{0.32\textwidth}
		\centering
		\includegraphics[width=\textwidth]{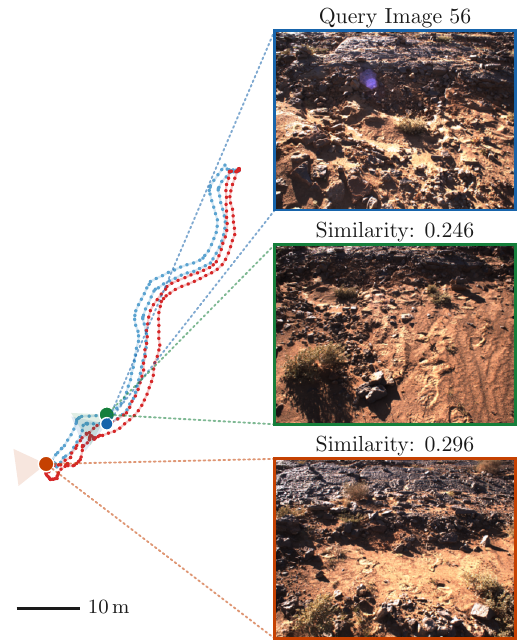}
		\caption{MADMAX A0 -- A2. Perceptual aliasing: The ``general look'' between the query (top) and best, but false positive match (bottom) is the same: a large boulder in the top half of the image, some grass below.}
	\end{subfigure}%
	\hfill
	\begin{subfigure}[t]{0.32\textwidth}
		\centering
		\includegraphics[width=\textwidth]{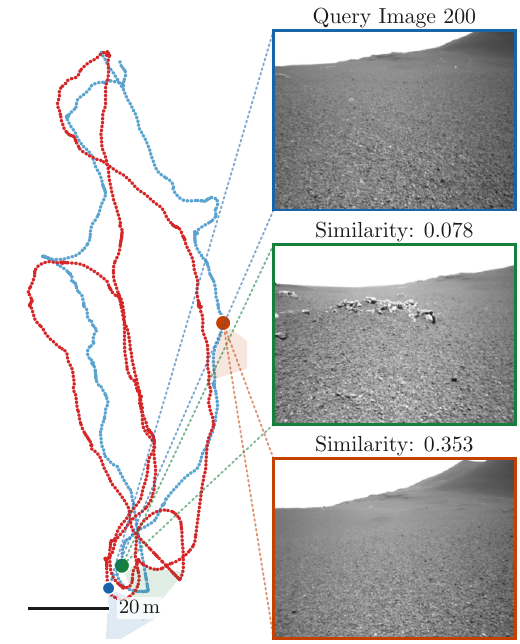}
		\caption{S3LI Etna traverse\_1 -- loops. Shared horizon: Best match database point is far away. The two images share the same horizon line, but an accurate relative pose estimation cannot be made -- the best 5DoF pose estimate being 39m off.}
		\label{fig:hard-queries:b}
	\end{subfigure}%
	\hfill
	\begin{subfigure}[t]{0.32\textwidth}
		\centering
		\includegraphics[width=\textwidth]{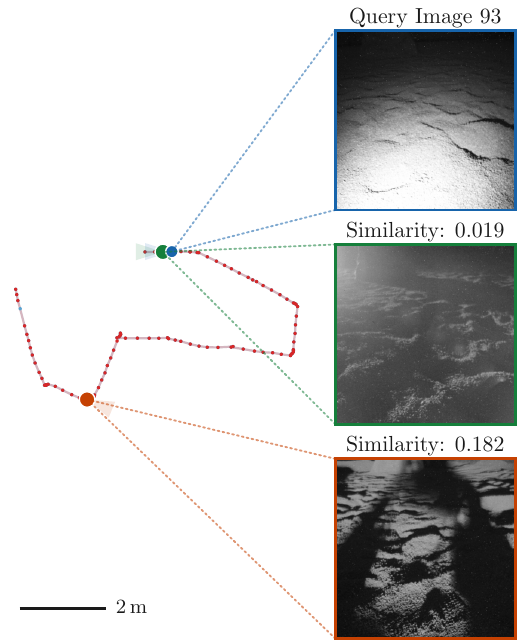}
		\caption{SPICE-HL³-SPAD dawn -- night. Query and positive database image are taken in close proximity. An image with similar lighting conditions is matched instead; 
		creating RPE translation- and rotation-errors of 4m and $87$° respectively.
		}
	\end{subfigure}
	\caption{Examples of hard queries in VPR-Space, illustrating the challenges of space analog domains. A query image (dark blue) has highest similarity to a false positive image (orange) outside its VPR-GT. The best correct image (green) inside the hard ground truth is displayed in the center. The margin between the similarities indicates the severity of the example.}
	\vspace{-0.3cm}
	\label{fig:hard-queries}
\end{figure*}

\input{paper_tables/auprc_table.tex}
\input{paper_tables/recall_table}

\section{Evaluation of SotA VPR approaches}
\label{sec:results}

\subsection{Quantative Analysis}
We evaluate MegaLoc \cite{berton2025megaloc} (MeLo), Bag-of-Queries \cite{ali2024boq} (BoQ), SALAD \cite{izquierdo2024optimal} (SLD), AnyLoc \cite{keetha2023anyloc} with the global vocabulary (AnyL), AnyLoc with the unstructured vocabulary, MixVPR \cite{ali2023mixvpr}, CosPlace \cite{berton2022rethinking} as well as DenseVLAD \cite{torii201524} (DenV), NetVLAD \cite{arandjelovic2016netvlad} and DINOv3 (CLS token) \cite{simeoni2025dinov3} as baselines. We standardize with the database statistics and L2-normalize all descriptors before constructing the similarity matrices. We report the Area under the Precision-Recall Curve (AUPRC) in Table \ref{tab:vpr-auprc} and Recall@$\{1,5,10\}$ as well as Recall@100p (recall at 100\% precision) in Table \ref{tab:vpr-recall}. The tables do not include AnyLoc with the unstructured vocabulary, MixVPR, CosPlace and DINOv3 as these methods didn't score highest on any of the datasets; fully detailed results will be made available on the benchmark website. 

The results indicate significant room for improvement for VPR methods on SAD. Averaged across all datasets, the best performing VPR approach achieves an AUPRC of 0.47 and recall@1 of 0.64  (with values of 0.18 and 0.38, respectively, for the most challenging dataset). Noticeable, in the worst case, a mode especially important for autonomous missions, R@100p is near zero across all methods.


Although being finetuned on urban data, Bag-of-Queries and MegaLoc demonstrate domain adaptation, as can be seen by their performance on the MADMAX A0 -- A2, MADMAX F0 -- F3, and SPICE-HL³ dawn -- noon pairs in Table \ref{tab:vpr-auprc}. Except for setups where outlines and shadows are quite similar, such as in SPICE-HL³'s dawn -- noon pairings, current VPR methods struggle when the difference in lighting becomes too large. Notably, DenseVLAD, a non deep learning method, performs best on SPICE-HL³-SPAD-dawn-noon. 

\subsection{Qualitative Analysis: What are hard queries that the SotA cannot solve yet?}

Three examples from the analysis of the BoQ results can be seen in Figure \ref{fig:hard-queries}. One common pattern can be seen in Figure \ref{fig:hard-queries:b}: in SAD and other domains with large and open spaces, images along a straight sequence with a shared horizon tend to have high similarity. This is especially relevant as image pairs of this failure mode may pass the geometric verification stage and hence produce a relative pose estimate with large error.

\section{Discussion}

Both RPEs and VPR methods rely on visual overlap; RPEs to determine a relative pose and VPR methods to produce an embedding. They may share failure modes, which would exaggerate performance in SAD. To this end, Figure \ref{fig:rep-advantages:b} can be interpreted as a possible limitation of our approach: Even though camera poses coincide, we do not require a match. However, we want to emphasize that our proposed VPR-GT is driven by the domain of interest and possible downstream application in a SLAM context. This results in a different notion of which matches should and should not be found. 

Further, our VPR-GT is constructed to go \emph{beyond} the capabilities of any one RPE, using a voting mechanism to increase robustness. 
As available RPEs improve or the expected characteristics of the downstream task change, RPE-based GT matrices may be updated. We refer to the GT used here as ``\emph{VPR-GT-RPE-0}''. If future work uses a modified GT, it should indicate this.

Although spatial verification stages and robust SLAM optimization back-ends reduce the influence of false-positive matchings~\cite{schubert2023visual}, under the proposed GT, they gain relevance again (and so do AUPRC and R@100p metrics that explicitly incorporate false positives). Due to the GT construction approach, false positive image pairs are \textit{known} to produce a large relative pose error. And due to the similar nature of RPE approaches and spatial verification stages, they are also likely to pass this verification stage. 

For brevity, we only evaluated holistic VPR descriptors. However, VPR-Space can also be used for evaluation of local-descriptor-based and hierarchical pipelines.

Finally, we want to make clear that the way we use RPEs for GT generation can not be used to solve the VPR problem itself, since the VPR-GT generation additionally requires GT pose information to evaluate the RPE results.

\section{Conclusion}

This work presented \emph{VPR-Space}, a novel VPR benchmark for space analog domains. We selected and curated relevant sequence pairs from existing SLAM Space datasets to represent relevant viewpoint- and illumination-changes. To connect the VPR evaluation to the downstream SLAM task, we proposed to leverage an ensemble of SotA relative pose estimators to compute the VPR ground truth, rewarding VPR retrieval only if a low error pose estimate can be made. The evaluation of SotA VPR methods shows significant headroom for improvement and the qualitative analysis showed concrete failure modes of the VPR SotA. Our hope is that this benchmark sets new goals for VPR developments.


\vspace{-0.1cm}
\appendix 

\textit{Sub-sampling routine:}
Due to different velocities and sampling rates of the sensor platforms, some sequences contain multiple images of the same place. We alleviate this issue by applying a rolling window approach for all but the SPICE-HL³ sequences which do not need this post-processing step. We briefly summarize the sub-sampling routine: The first image of the sequence is added to the result list and becomes the previously added image. Then, for every following image, we check whether the associated $\Delta$ between the previous and current position exceeds 1m or the $\Delta$ in camera-yaw exceeds 10°. In which case we add the current image to the result list and the current image becomes the previous added image.

\input{paper_tables/hyperparam_table}

\section*{Acknowledgments and AI Statement}

We'd like to thank our colleagues Ida, Nick, Janine and Lena for proofreading. We'd like to thank Leon and Lorenz for helping with data annotation.\\

We used Claude Code to generate code to access datasets and existing approaches (RPEs as well as VPR), to support initial literature review as well as generation of figures, and to improve formal language.

\bibliographystyle{plain}
\bibliography{resources}

\end{document}

%% file: paper_tables/auprc_table.tex
\begin{table}[t]
  \centering
  \footnotesize
  \setlength{\tabcolsep}{3pt}
  \caption{AUPRC of SotA VPR methods on all sequence comparisons. Best per row in bold, runner-up underlined.}
  \label{tab:vpr-auprc}
  \begin{tabular}{@{}clrrrrr@{}}
    \toprule
    \shortstack{VPR\\chall.} & DB/Q pair & \shortstack{MeLo\\\cite{berton2025megaloc}} & \shortstack{BoQ\\\cite{ali2024boq}} & \shortstack{SLD\\\cite{izquierdo2024optimal}} & \shortstack{AnyL\\\cite{keetha2023anyloc}} & \shortstack{DenV\\\cite{torii201524}} \\
    \midrule
    \multirow{10}{*}{\rotatebox[origin=c]{90}{\shortstack{Viewpoint\\change}}} & MM A0 -- A2 & 0.67 & \textbf{0.70} & 0.65 & 0.68 & \underline{0.69} \\
     & MM A2 -- A4 & 0.36 & \textbf{0.41} & \underline{0.39} & 0.29 & 0.32 \\
     & MM C0 -- C1 & \underline{0.51} & \textbf{0.52} & 0.41 & 0.33 & 0.31 \\
     & MM E0 -- E1 & \textbf{0.33} & \underline{0.30} & 0.23 & 0.26 & \underline{0.30} \\
     & MM F0 -- F3 & \textbf{0.79} & \underline{0.71} & 0.68 & 0.63 & 0.57 \\
     & MM G0 -- G2 & \textbf{0.57} & 0.49 & 0.48 & \underline{0.50} & 0.41 \\
     & S3E crater -- crater\_inout & \textbf{0.52} & 0.33 & 0.32 & 0.18 & \underline{0.35} \\
     & S3E traverse\_1 -- loops & \textbf{0.20} & \underline{0.17} & 0.10 & 0.08 & 0.07 \\
     & LL 9 -- 10 & \textbf{0.62} & \underline{0.56} & 0.36 & 0.44 & 0.41 \\
     & LL 15 -- 17 & 0.48 & \underline{0.49} & \textbf{0.51} & 0.40 & 0.34 \\
    \midrule
    \multirow{6}{*}{\rotatebox[origin=c]{90}{\shortstack{Illumination\\change}}} & HL$^3$ SPAD dawn -- night & 0.18 & \underline{0.21} & 0.13 & \textbf{0.23} & 0.02 \\
     & HL$^3$ SPAD dawn -- noon & 0.76 & \underline{0.79} & 0.69 & 0.66 & \textbf{0.80} \\
     & HL$^3$ SPAD noon -- night & \underline{0.21} & \textbf{0.35} & 0.19 & \textbf{0.35} & 0.05 \\
     & HL$^3$ FLIR dawn -- night & \underline{0.33} & \textbf{0.35} & 0.27 & 0.32 & 0.17 \\
     & HL$^3$ FLIR dawn -- noon & \underline{0.70} & \textbf{0.72} & 0.60 & 0.65 & 0.67 \\
     & HL$^3$ FLIR noon -- night & 0.28 & \textbf{0.41} & 0.31 & \underline{0.33} & 0.12 \\
    \midrule
    \multirow{3}{*}{\rotatebox[origin=c]{90}{}} & Worst case & \textbf{0.18} & \underline{0.17} & 0.10 & 0.08 & 0.02 \\
     & Avg. mean & \textbf{0.47} & \textbf{0.47} & 0.39 & \underline{0.40} & 0.35 \\
     & Avg. std & $\pm$0.20 & $\pm$0.19 & $\pm$0.19 & $\pm$0.18 & $\pm$0.24 \\
    \bottomrule
  \end{tabular}
\vspace{-0.5cm}
\end{table}

%% file: paper_tables/recall_table.tex
\begin{table}[t]
  \centering
  \footnotesize
  \setlength{\tabcolsep}{7pt}
  \caption{Recall@k and recall@100\% precision of SotA VPR methods. Best per row in bold, runner-up underlined}
  \label{tab:vpr-recall}
  \begin{tabular}{@{}clrrrr@{}}
    \toprule
    Case & Method & R@1 & R@5 & R@10 & R@100p \\
    \midrule
    \multirow{5}{*}{\rotatebox[origin=c]{90}{\shortstack{Worst\\case}}} & MeLo & 0.24 & 0.46 & 0.53 & 0.00 \\
     & BoQ & \textbf{0.38} & \textbf{0.59} & \textbf{0.69} & 0.00 \\
     & SLD & 0.25 & 0.56 & 0.66 & 0.00 \\
     & AnyL & 0.31 & 0.54 & 0.64 & \textbf{0.001} \\
     & DenV & 0.06 & 0.23 & 0.33 & 0.00 \\
    \midrule
    \multirow{5}{*}{\rotatebox[origin=c]{90}{Avg.\ $\pm$ std}} & MeLo & 0.59{\scriptsize$\pm$0.25} & 0.75{\scriptsize$\pm$0.19} & 0.81{\scriptsize$\pm$0.15} & 0.07{\scriptsize$\pm$0.07} \\
     & BoQ & \textbf{0.64}{\scriptsize$\pm$0.20} & \textbf{0.78}{\scriptsize$\pm$0.14} & \textbf{0.84}{\scriptsize$\pm$0.11} & \textbf{0.08}{\scriptsize$\pm$0.08} \\
     & SLD & 0.59{\scriptsize$\pm$0.21} & 0.77{\scriptsize$\pm$0.14} & \textbf{0.84}{\scriptsize$\pm$0.12} & 0.05{\scriptsize$\pm$0.07} \\
     & AnyL & 0.56{\scriptsize$\pm$0.21} & 0.73{\scriptsize$\pm$0.16} & 0.81{\scriptsize$\pm$0.13} & 0.04{\scriptsize$\pm$0.04} \\
     & DenV & 0.52{\scriptsize$\pm$0.27} & 0.68{\scriptsize$\pm$0.24} & 0.76{\scriptsize$\pm$0.20} & 0.05{\scriptsize$\pm$0.05} \\
    \bottomrule
  \end{tabular}
	\vspace{-0.4cm}
\end{table}

%% file: paper_tables/hyperparam_table.tex
\begin{table}[htb]
  \centering
  \footnotesize
  \setlength{\tabcolsep}{3pt}
  \begin{tabular*}{\linewidth}{@{\extracolsep{\fill}}lrrrr@{}}
    \toprule
     & \multicolumn{2}{c}{\# Images} & \% queries & \% of pos. \\
    \cmidrule(lr){2-3}
    DB/Q pair & DB & Q & w/ match & matches \\
    \midrule
    MM A0 -- A2 & 171 & 180 & 92.8 & 8.55 \\
    MM A2 -- A4 & 180 & 412 & 20.9 & 0.74 \\
    MM C0 -- C1 & 502 & 463 & 78.6 & 1.89 \\
    MM E0 -- E1 & 517 & 507 & 43.2 & 0.62 \\
    MM F0 -- F3 & 117 & 229 & 89.1 & 5.14 \\
    MM G0 -- G2 & 258 & 290 & 51.7 & 1.75 \\
    S3E crater -- crater\_inout & 1129 & 1683 & 18.2 & 0.15 \\
    S3E traverse\_1 -- loops & 485 & 780 & 36.8 & 0.40 \\
    LL 9 -- 10 & 119 & 183 & 71.0 & 3.19 \\
    LL 15 -- 17 & 118 & 54 & 27.8 & 0.97 \\
    HL$^3$ SPAD dawn -- night & 98 & 97 & 76.3 & 1.25 \\
    HL$^3$ SPAD dawn -- noon & 98 & 97 & 100.0 & 6.00 \\
    HL$^3$ SPAD noon -- night & 97 & 97 & 72.2 & 1.74 \\
    HL$^3$ FLIR dawn -- night & 99 & 97 & 78.4 & 2.55 \\
    HL$^3$ FLIR dawn -- noon & 99 & 99 & 96.0 & 5.93 \\
    HL$^3$ FLIR noon -- night & 99 & 97 & 78.4 & 2.72 \\
    \bottomrule
  \end{tabular*}
  \caption{Additional dataset statistics. For each database/query pair we report the number of images, the percentage of queries which have at least one match (\% queries w/ match) and the percentage of positive matches (\% of pos. matches) given the proposed VPR-GT-RPE-0.}
  \label{tab:vpr-data}
  \vspace{-0.3cm}
\end{table}

\begin{table}[htb]
  \centering
  \footnotesize
  \setlength{\tabcolsep}{3pt}
  \begin{tabular*}{\linewidth}{@{\extracolsep{\fill}}lcccccccc@{}}
    \toprule
     & \multicolumn{2}{c}{\textbf{MADMAX}} & \multicolumn{2}{c}{\textbf{S3LI Etna}} & \multicolumn{2}{c}{\textbf{LunarLoc}} & \multicolumn{2}{c}{\textbf{SPICE-HL$^3$}} \\
    \cmidrule(lr){2-3} \cmidrule(lr){4-5} \cmidrule(lr){6-7} \cmidrule(lr){8-9}
     & soft & hard & soft & hard & soft & hard & soft & hard \\
    \midrule
    $d_{\max}$ [m] & 35 & 35 & 35 & 35 & 35 & 35 & 1 & 1 \\
    $\tau_t$ [m] & 4 & 3 & 4 & 3 & 4 & 3 & 0.4 & 0.2 \\
    $\tau_R$ [$^\circ$] & 30 & 20 & 30 & 20 & 30 & 20 & 20 & 10 \\
    $n$ & 2 & 4 & 2 & 4 & 2 & 4 & 1 & 1 \\
    \bottomrule
  \end{tabular*}
  \caption{Hyperparameters used for VPR-GT-RPE-0.}
  \label{tab:vpr-gt}
  \vspace{-0.5cm}
\end{table}